\documentclass[preprint,11pt]{elsarticle}
\usepackage{epstopdf}
\usepackage{caption}
\usepackage{setspace}
\usepackage{amssymb}
\usepackage{multirow}
\usepackage{amsthm}
\usepackage{booktabs}
\usepackage{amsmath}
\usepackage{amssymb}
\usepackage{mathrsfs}
\usepackage{longtable}
\usepackage{lscape}
\usepackage{url}
\usepackage{tabularx}
\usepackage{epsfig}
\usepackage{colortbl}
\usepackage{subfigure}
\usepackage{tikz,mathpazo}
\usepackage{graphicx}
\usepackage{natbib}
\usepackage[noend]{algpseudocode}
\usepackage{algorithm}
\usepackage{algorithmicx}

\newtheorem{example}{Example}[section]

\newtheorem{definition}{Definition}[section]
\usetikzlibrary{shapes.geometric, arrows}
\biboptions{numbers,sort&compress}

\begin{document}
\begin{frontmatter}
\title{Fortified quantum mass function utilizing ordinal
	pictorial check based on time interval analysis and
	expertise}
\author[address1]{Yuanpeng He}
\author[address1]{Yipeng Xiao \corref{label}}  
\author[address1]{Bo Gou \corref{label}} 
\author[address1]{Fuyuan Xiao\corref{label1}}
\address[address1]{School of Computer and Information Science, Southwest University, Chongqing, 400715, China}
\cortext[label1]{Corresponding author: Fuyuan Xiao, School of
	Computer and Information Science, Southwest University, Chongqing,
	400715, China. Email address: xiaofuyuan@swu.edu.cn,
	doctorxiaofy@hotmail.com.}
\cortext[label]{The contributions of the authors are equal.}
\begin{abstract}
Information management has enter a completely new era, quantum era. However, there exists a lack of sufficient theory to extract truly useful quantum information and transfer it to a form which is intuitive and straightforward for decision making. Therefore, based on the quantum model of mass function, a fortified dual check system is proposed to ensure the judgment generated retains enough high accuracy. Moreover, considering the situations in real life, everything takes place in an observable time interval, then the concept of time interval is introduced into the frame of the check system. The proposed model is very helpful in disposing uncertain quantum information in this paper. And some applications are provided to verify the rationality and correctness of the proposed method.
\end{abstract}
\end{frontmatter}
\section{Introduction}
In recent years, the management of information has become a very hot field. A lot of relevant works have been completed to provided different kinds of method to properly handle information offered which promotes the development of information industry. The representatives of the corresponding theories are soft theory \cite{Fei2019Evidence,Alcantud2019An,fei2019pythagorean,feng2016soft,Song2019POWA}, $Z$-numbers \cite{Jiang2019Znetwork,tian2020zslf,kang2019deriveknowledge,zadeh2011note}, $D$-numbers \cite{IJISTUDNumbers, liu2020anextended,LiubyDFMEA,deng2019evaluating,mo2021swot}, fuzzy theory \cite{Xiao2019Distancemeasure,fei2020mcdmPFS,song2019divergencebased,zhanglimao2019}, Dempster-Shafer evidence theory \cite{deng2020negationINS, yager2019generalized, Luo2020vectorIJIS,Luo2020negation,Dempster1967Upper} and some other mixed theories \cite{Xiao2020GIQ,Xiao2020Novel,jiang2018Correlation}. And the effectiveness of these theories are verified in many practical applications, like risk evaluation \cite{fu2020evidential,pan2020multi,8926527}, pattern classification \cite{liu2020evidence}, optimization \cite{xu2016optimal,meng2018fluid,cheong2019hybrid,yang2019networkPhysica} and decision making \cite{DBLP:journals/ijar/Smets05,Harish2019NovelAggregation,Zhan2020Three,liu2020ahesitant}. Moreover, due to the rapid progress of quantum computing, some researchers come up with the idea that traditional information management can be transferred to the level of quantum. Some meaningful works about the topic are complex mass function \cite{Xiao2020CEQD,garg2019some,Xiao2019complexmassfunction,Xiao2020maximum,Xiao2020generalizedbelieffunction} and quantum information theory \cite{lai2020parrondo,dengentropyentanglement,Dai2020interferenceQLBN,Gao2019quantummodel}. In this paper, the proposed method is based on the quantum model of mass function \cite{Gao2019quantummodel}. In order to avoid the deviation which may caused by the original quantum evidences, a dual check system is designed to ensure the authenticity of the original judgments which utilizes the concept of $Z$-number \cite{zadeh2011note}. Besides, because of the introduction of the time interval, a specially devised rule is proposed to appropriately decide the importance of different relationships of incidents, which is a kind of expert system under some restrictions. The contributions of the proposed method can be listed as:
\begin{enumerate}[(1)]
	\item The second dual check system can help avoid the deviation produced by the original evidences to help provide more effective results.
	\item Introduction of time interval make the frame of discernment more adaptive to the real life.
	\item The fortified quantum mass function is able to produce intuitive and reasonable judgments about current situations compared with traditional rule of combination.
\end{enumerate}

The rest of this paper is organized as follows. The part of preliminaries generally introduces some basic concepts of the proposed method. Then, the following section provides details of the fortified mass function. Besides, in the part of application, two applications are provided to prove the superiority and validity of the method in this paper. In the last part, some conclusive opinions are made to summarise advantages of the method proposed in this paper.

\section{Preliminaries}
\label{Preliminaries}
In this part, some related concepts are briefly introduced. And there exists lots of works which solves problems in relative fields \cite{Deng2020ScienceChina,LiuzyINS15757,Deng2020InformationVolume,yang2019network}.
\subsection{Quantum model of mass function \cite{Gao2019quantummodel}}
\begin{definition}(Quantum mass function)\end{definition}

Assume there exists a quantum frame of discernment, in which a quantum mass function $Q$ can be defined as: 

\begin{equation}
Q(|A\rangle)=\psi e^{i\theta}
\end{equation}
The quantum mass function is also named as quantum basic probability assignment (QBPA), which is a mapping of $Q$ from $0$ to $1$ and the properties it satisfies are given as:
\begin{equation}
Q(\phi)=0
\end{equation}
\begin{equation}
\sum_{|A\rangle \subseteq \Omega }|Q(|A\rangle)|=1
\end{equation}
The value of $|Q(|A\rangle)|$ equals to $\psi^{2}$, which is regarded as the  degree of belief to $|A\rangle$. Besides, the phase angle of $|A\rangle$ is represented by $\theta$.

\textbf {Remark 1:} The quantum mass function can degenerate to classical mass function when the phase angle of quantum mass function equals $0^{\circ}$.

\textbf {Remark 2:} Additivity is not  satisfied in the quantum mass function, which is defined as:

\begin{equation}
|Q(A\rangle)+Q(B\rangle)|\neq|Q(A\rangle)|+|Q(B\rangle)|
\end{equation}

\subsection{Z-numbers \cite{zadeh2011note}}
Z-number is a relatively new instrument to measure the reliability of the information. The credibility of information gotten plays an significant part in information extraction and disposal.

\begin{definition}(Z-numbers)\end{definition}

A Z-number is a kind of group of fuzzy numbers, which is expressed as $Z=(A, B)$. A Z-number is bound up with $X$, which is an indeterminate variable with real value. The first constituent part $A$  is sort of constraint of $Z$. Besides, the second component  $B$ is an instrument of measure of the degree of reliability of the first constituent part $A$.

\subsection{Picture fuzzy set (PFS)}
A picture fuzzy set $A$ on the finite universe of discourse $X$ whose mathematical form is represented as:

\begin{equation}
A=\{\langle x, u_{A}(x), \epsilon_{A}(x),v_{A}(x) \rangle | x \in X\} 
\end{equation}

with the condition:

\begin{equation}
\begin{aligned}
	&u_{A}(x):X\rightarrow [0,1] 
	\\&\epsilon_{A}(x):X\rightarrow [0,1]
	\\&v_{A}(x):X\rightarrow [0,1]
	\\&0 \leq u_{A}(x) +\epsilon_{A}(x)+v_{A}(x) \leq 1
\end{aligned}
\end{equation}

The $u_{A}(x) $ represents the membership degree of $x \in X$. Similarly, $v_{A}(x)$ is the non-membership degree of $x \in X$. Besides, $\epsilon_{A}(x)$ is called the degree of hesitancy  $x \in X$ .

For a PFS $A$ in $X$, a refusal function of $x \in X$ which represents the degree of refusal  can be defined as:

\begin{equation}
\S_{A}(x)=1-(u_{A}(x)+\epsilon_{A}(x)+v_{A}(x)),\forall x \in X
\end{equation} 

Largely, picture fuzzy set is appropriate for these situations when decision makers face their opinions involving with their determination making as follows: support(yes), neutrality(hesitancy), oppose(no) and refusal. For example, four probable circumstances people may encounter in voting are: "vote for", "abstain", "vote against" and "refusal of the voting". The intentions of group "vote for" and "vote against" are apparent. Group "abstain" means that voters hesitate between "vote for "and "vote against". As for group "refusal of voting", which denotes either invalid voting papers or abstention.

\subsection{Ordinal Quantum Frame of Discernment (OQFD)}
The ordinal quantum frame of discernment is a set whose propositions are relational with a certain order, which is defined as:

\begin{equation}
\Theta_{ordinal}=\{M_{1},M_{2},...,M_{n}\}
\end{equation}

The sequence of proposition is expressed by subscripts. The propositions in the ordinal frame of discernment satisfy the following characters:

\begin{itemize}
\item For any element with subscript $i$, it is supposed to be confirmed before the ones with subscripts $i+n(n \geq 1)$.
\item The definition of proposition in an ordinal frame of discernment is the same as the ones defined in the traditional quantum frame of discernment besides order of elements.
\item The degree of uncertainty of whole evidence system can be  further ascertained.
\end{itemize}

\section{Proposed method}

In the traditional ordinal quantum frame of discernment, propositions are associated in a certain order. The subscript $i$ represents the order of them. Traditional ordinal quantum frame of discernment takes the influence of propositions which take place before its later ones preliminarily .
However, previous frame of discernment does not take the detail of true situation so alleged ordinal relationship is opinionated and does not accord with the actual condition. As a result, the consequence of prediction based on traditional ordinal quantum frame of discernment do not possess enough accuracy and could not reflect the true circumstance. Hence, in order to describe the connection between sequential  propositions appropriately, time interval is added into the $OQFD$.

\subsection{Ordinal Quantum Frame of Discernment with Time Interval (OQFDTI)}

The ordinal quantum frame of discernment with time interval is a set whose elements are related in a certain order, which is defined as:

\begin{equation}
\Theta_{OQFDTI}=\left\{
\begin{array}{rcl}
	M_{1}^{\alpha}\ \dashrightarrow M_{2}^{\beta}& & Time \in [\vartheta,\upsilon]\\
	M_{2}^{\beta}\ \rightarrowtail M_{3}^{\gamma}& & Time \in [\kappa,\xi]\\
	M_{3}^{\gamma}\ \twoheadrightarrow M_{4}^{\delta}& & Time \in [\rho,\varsigma]\\
	.&&.\\
	.&&.\\
	.&&.\\
	M_{n-1}^{\zeta}\ \Rrightarrow M_{n}^{\varpi}& & Time \in [\tau,\omega]\\
\end{array} \right.
\end{equation}

The sequence of propositions is denoted by subscripts. The order of propositions is stipulated by time sequence, the propositions which possess smaller subscripts are proposed earlier. Superscripts represent the time when the propositions are raised. Different time intervals between sequential two propositions contained in the frame of discernment signify the degree of closeness of connection. Time interval is shorter, the relation between two propositions is closer, which means that the influence of the antecedent proposition on the  subsequent proposition is bigger. In $OQFDTI$, time interval given by sensors and the order of incident occurred ae taken into consideration which make the forecast reflect the essence of the objects further. As a result, the frame mentioned above possesses more accuracy and rationality.

The following properties of elements in $OQFDTI$ satisfied are listed:

\begin{itemize}
\item Diverse styles of arrows represent different degree of relation between any two sequential propositions in corresponding time interval.
\item The degree of uncertainty of the whole quantum system can be further ascertained in the course of determining one more proposition.
\item For any element with a subscripts $i$, it is supposed to be confirmed before the proposition with subscripts $i+n$ ($n \geq 1$).
\item The definition of proposition in an ordinal frame of discernment is exactly the same as the ones defined in traditional ordinal quantum frame of discernment except time interval.
\end{itemize}

\subsection{Time quantitative rule}
Since the $OQFDTI$ is ordinal, there exists a decisive relationship between the uncertainty of evidence system and the time intervals of two sequential propositions. First of all, in $OQFDTI$, the propositions which occur in the first place are considered to have a crucial effect on the propositions which occur after them. Besides, $OQFDTI$ mentioned, different time intervals between sequential two propositions contained in the frame of discernment signify the degree of the closeness of connection of them. Because the multifarious time intervals given by senors are fluctuating, how to confirm three disparate degree of connection of every time interval\\($	\Upsilon_{strong} ,	\Upsilon_{moderate},\Upsilon_{weak}$) should be taken into account which signify strong, moderate and weak connection of two sequential propositions in according time intervals. On the basis of the definition of $OQFDTI$, the detailed process of getting modified values of each proposition is defined as:

(1)According to the weights denoted by $Wgt$ given by sensors, the modified weights of  proposition $i$ can be expressed specifically as:

	\begin{equation}
		Weights^{Time}_{i}=\left\{
		\begin{array}{rcl}
			Wgt \times \Upsilon_{strong} & &\varkappa \leqslant Time < \frac{\varkappa+\digamma}{2}     \\
			Wgt \times \Upsilon_{moderate} & & Time=\frac{\varkappa+\digamma}{2} \\
			Wgt \times \Upsilon_{weak} & & \frac{\varkappa+\digamma}{2} <Time \leqslant \digamma\\
			
		\end{array} \right.
		Time \in [\varkappa,\digamma]
	\end{equation}

\textbf{Note: }The weight of the first proposition is $1$.

(2)For $OQFDTI$, mass of proposition with subscript $i$ is assemble into a group with proposition whose position is  $i-1$ except $i=1$. Original mass of every proposition is denoted by $Mass_{i}$ and the process of obtaining modified intermediate values of each proposition of an evidence is defined as:

\begin{tiny}
\begin{equation}
	\begin{bmatrix}
		Value_{1}^{inter}&Value_{2}^{inter}&...&Value_{i}^{inter}&
	\end{bmatrix}=\begin{bmatrix}
		Mass_{1}&	Mass_{2}&...&	Mass_{i}&
	\end{bmatrix} \times \begin{bmatrix}
		Weights^{Time}_{1}\\
		&\ddots\\
		&\ & Weights^{Time}_{i}
	\end{bmatrix}
\end{equation}
\end{tiny}

\textbf{Note: }For mass of each proposition with subscript $h$ must be confirmed earlier than the ones with subscript $i(i>h)$.

\textbf{Example: }Assume there exists three propositions $\{	M_{1}^{\alpha},M_{2}^{\beta},M_{3}^{\gamma}\}$. When you try to ascertain  $Weights^{Time}_{3}$ for $M_{3}^{\gamma}$ in the corresponding interval $M_{2}^{\beta} \rightarrowtail M_{3}^{\gamma}$ $Time \in [\kappa,\xi]$, the $Weights^{Time}_{2}$ for $M_{2}^{\beta}$ must be determined in terms of $M_{1}^{\alpha} $ in the interval $M_{1}^{\alpha} \dashrightarrow M_{2}^{\beta}$ $Time \in [\vartheta,\upsilon]$.

(3)The step of normalization of modified intermediate value of proposition $i$ is defined as:

\begin{equation}
\begin{bmatrix}
	Value_{1}^{inter}&Value_{2}^{inter}&...&Value_{i}^{inter}&
\end{bmatrix}
\rightarrow
\begin{bmatrix}
	Value_{1}^{nor}&Value_{2}^{nor}&...&Value_{i}^{nor}&
\end{bmatrix}
\end{equation}
\textbf{Note: }The step of normalization is obtaining the quotient respectively of each intermediate value over the sum of all intermediate values.

\subsection{$TDQBF$:TWO-DIMENSIONAL QUANTUM BELIEF FUNCTION }
\begin{definition}A $TDQBF$, M=(m$_1$,m$_2$). m$_1$, m$_2$ are also $QBPAs$ and $m_2$ is a measure of reliability of m$_1$.\end{definition}
The elements of $m_1$ and $m_2$ are consist of quantum probability assignment. The quantum frame of discernment of $m_2$ can be represented by $\Theta$=$\left\{Y,N,H,R\right\}$. 

Assume there is a voting, the tickets for support, opposition, waiver neutral(hesitating), abstention are denoted by $\left\{Y\right\}$, $\left\{N\right\}$, {$\left\{H\right\}$} and $\left\{R\right\}$ respectively.
\subsection{$DHDF$:Dynamic Hesitation Distribution Formula}
Assume there exists three $QBPAs$, which are $m_2(Y)$, $m_2(N)$ and $m_2(H)$, $DHDF$ can be shown as follow:
\begin{equation}
m_Y(H)=\sqrt{\frac{|m_2(Y)|^2}{(|m_2(Y)|+|m_2(N)|+2|m_2(Y)||m_2(N)|)^2}} \times m_2(H)
\end{equation}
\begin{equation}
m_N(H)=\sqrt{\frac{|m_2(N)|^2}{(|m_2(Y)|+|m_2(N)|+2|m_2(Y)||m_2(N)|)^2}} \times m_2(H)
\end{equation}

Because the degree of uncertainty should be reduced to make the results more deterministic, a part of hesitance should be distributed to corresponding propositions which means a part of mass of $m_2(H)$ should be allocated to $m_2(Y)$ and $m_2(N)$. And the mass assigned to $m_2(Y)$ and $m_2(N)$ are represented by $m_Y(H)$ and $m_N(N)$ respectively.

\subsection{$QPDR$:Quantum Pignistic Distribution rule }
Assume there exists three $QBPAs$, which are $m_2(Y)$, $m_2(N)$ and $m(H)$, $QPDR$ can be defined as:
\begin{equation}
m_Z(Y)=\sqrt{\frac{|m_2(Y)|^2}{|m_2(Y)|^2+|m_2(N)|^2}}\times m_Y(H)+ m_2(Y)
\end{equation} 
\begin{equation}
m_Z(N)=\sqrt{\frac{|m_2(N)|^2}{|m_(Y)|^2+|m_2(N)|^2}}\times m_N(H)+m_2(N)
\end{equation}

where the results of two $QBPAs$ after distribution are expressed as $m_Z(Y) $ and $m_Z(N)$. $m_Y(H)$ and $m_N(H)$  represent the mass of $m_2(H)$ distributed to $m_2(Y)$ and $m_2(N)$ respectively. The variables $\sqrt{\frac{|m_2(Y)|^2}{|m_2(Y)|^2+|m_2(N)|^2}}$ and $\sqrt{\frac{|m_2(N)|^2}{|m_(Y)|^2+|m_2(N)|^2}}$ aim to solve the problem about the loss when the transferring the quantum into the form of classic probability. The method in this part decrease degree of uncertainty and conform to the actuality and reflect the internal connections of thighs.

\textbf{Example3.1}
Assume there are three $QBPAs$, which are be defined as:
$m_2(Y)$=$\frac{\sqrt{2}}{4}$+$\frac{\sqrt{2}}{4}$i,
$m_2(N)$=$\frac{\sqrt{2}}{4}$+$\frac{\sqrt{2}}{4}$i, and
$m_2(H)$=$\frac{1}{2}$ +$\frac{1}{2}$i. 

If the mass of $m_2(H)$ assigns to $m_2(Y)$ and $m_2(N)$ totally. The value of $m_2(H)$ distributed to to $m_2(Y)$ and $m_2(N)$ by classic method are:
$m_y(H)$=$\frac{1}{4}$ +$\frac{1}{4}$i, $m_n(H)$=$\frac{1}{4}$+$\frac{1}{4}$i. And the sum of classic probability of $m_y(H)$ and $m_n(H)$ are $\frac{1}{4}$.\\

$m_y(H)$=$\frac{|m_2(Y)|^2}{|m_2(Y)|^2+|m_2(N)|^2}$$\times m_2(H)$=$\frac{1}{4}$ +$\frac{1}{4}$i

$P1$=$|m_y(H)|^2$=$\frac{1}{8}$

$m_n(H)$=$\frac{|m_2(N)|^2}{|m_2(Y)|^2+|m_2(N)|^2}$$\times m_2(H)$=$\frac{1}{4}$ +$\frac{1}{4}$i

$P2$=$|m_n(H)|^2$=$\frac{1}{8}$

$classic\ \ probability$=$P1$+$P2$=$\frac{1}{4}$\\

The value obtained by formula $\sqrt{\frac{|m_2(Y)|^2}{|m_2(Y)|^2+|m_2(N)|^2}}$ and $\sqrt{\frac{|m_2(N)|^2}{|m_(Y)|^2+|m_2(N)|^2}}$ are:
$m_y(H)$=$\frac{\sqrt{2}}{4}$ +$\frac{\sqrt{2}}{4}$i, $m_n(H)$=$\frac{\sqrt{2}}{4}$ +$\frac{\sqrt{2}}{4}$i. And the sum of classic probability of $m_y(H)$ and $m_n(H)$ are $\frac{1}{2}$.\\

$m_y(H)$=$\frac{|m_2(Y)|^2}{|m_2(Y)|^2+|m_2(N)|^2}$$\times m_2(H)$=$\frac{\sqrt{2}}{4}$ +$\frac{\sqrt{2}}{4}$i

$P1$=$|m_y(H)|^2$=$\frac{1}{4}$

$m_n(H)$=$\frac{|m_2(N)|^2}{|m_2(Y)|^2+|m_2(N)|^2}$$\times m_2(H)$=$\frac{\sqrt{2}}{4}$ +$\frac{\sqrt{2}}{4}$i

$P2$=$|m_n(H)|^2$=$\frac{1}{4}$

$classic\ \ probability$=$P1$+$P2$=$\frac{1}{2}$\\

The classic probability of $m_2(H)$=$\frac{1}{2}$ +$\frac{1}{2}$i is $\frac{1}{2}$. So, the loss of quantum probability by classic methods can be solved by the formula in this part.\\

$classic\ \ probability$=$|m_2(H)|^2$=$\frac{1}{2}$\\

\textbf{Example3.2}
Assume there are three $QBPAs$, which are $m_2(Y)$=$\frac{2\sqrt{2}}{5}$+$\frac{2\sqrt{2}}{5}$i,
$m_2(N)$=$\frac{2\sqrt{2}}{5}$+$\frac{2\sqrt{2}}{5}$i,
and $m_2(H)$=$\frac{1}{2}$ +$\frac{1}{2}$i. 

The results are listed as follows:
$m_Z(Y)$=$\frac{61\sqrt{2}}{165}$ + $\frac{61\sqrt{2}}{165}$i
,$m_Z(N)$=$\frac{61\sqrt{2}}{165}$  + $\frac{61\sqrt{2}}{165}$i\\

$m_Z(Y)$=$\sqrt{\frac{|m_2(Y)|^2}{|m_2(Y)|^2+|m_2(N)|^2}}\times m_Y(H)+ m_2(Y)$=$\frac{61\sqrt{2}}{165}$ + $\frac{61\sqrt{2}}{165}$i

$m_Z(N)$=$\sqrt{\frac{|m_2(N)|^2}{|m_2(Y)|^2+|m_2(N)|^2}}\times m_Y(H)+ m_2(N)$=$\frac{61\sqrt{2}}{165}$ + $\frac{61\sqrt{2}}{165}$i\\

\subsection{The combination of $TDQBF$}
$TDQBF$ can be defined as:
\begin{equation}
\begin{split}
	m_X(x_i)=m_1(\left\{x_i\right\})\times m_Z(\left\{Y\right\})+(1-m_1(\left\{x_i\right\}))\times m_Z(\left\{N\right\})\\ 
	+\sum_{x_i\subseteq x_\alpha} m_1(x_\alpha) \times  m_Z(\left\{Y\right\}) +\sum_{x_i\nsubseteq x_\beta} m_1(x_\beta) \times 	m_Z(\left\{N\right\})  
	\quad \forall x_i\subset 2^\Theta 
\end{split}
\label{TDQBF1}
\end{equation}
\begin{equation}
m_X(A_i)=m_1(\left\{A_i\right\})\times m_Z(\left\{Y\right\})+m_1(\Theta) \times m_Z(\left\{Y\right\}) \quad \forall A_i\subset2^\Theta
\label{TDQBF2}
\end{equation}
\begin{equation}
m_X(\Theta)=m_1(\Theta) \times m_2(\left\{R\right\})
\label{TDQBF3}
\end{equation}

where $x_i$ is a single subset of $2^\Theta$ , $A_i$ is multisubset of $2^\Theta$, $i=1,2,3,4,...,n$. $x_i$ which is a single subset included in mutisubset $A_\alpha$ and not included in multisubset $A_\beta$. After distributing $m_2(\left\{H\right\})$ based on $QPDR$ to $m_2(\left\{Y\right\})$ and $m_2(\left\{N\right\})$, the two elements are denoted by $m_Z(\left\{Y\right\})$ and $m_Z(\left\{N\right\})$.

In equation \ref{TDQBF1}, the mass of $m_Z(\left\{Y\right\})$ is distributed to the $m_1(\left\{x_i\right\})$ proportionally. Besides, mass of $m_Z(\left\{N\right\})$ is distributed to the reverses of single subsets proportionally. Because single proposition is a part of  multiple propositions,  $\sum_{x_i\subseteq x_\alpha} m_1(x_\alpha) \times  m_Z(\left\{Y\right\})$ takes this effect of this situation into the formula, which means the mass of multiple propositions is divided into single proposition. In the same way, the case that single propositions calculated are not an element of multiple propositions is reflected in the part of formula, $\sum_{x_i\nsubseteq x_\beta} m_1(x_\beta) \times  m_Z(\left\{N\right\})$.

In equation \ref{TDQBF2}, the mass of $m_Z(\left\{Y\right\})$ is distributed to the $m_1(\left\{A_i\right\})$. Since the multisubset $A_i$ is also included in $\Theta$, $m_1(\Theta) \times m_Z(\left\{Y\right\})$ is utilized for dividing multisubset into universal set.

In equation \ref{TDQBF3}, Because $\Theta$ is symbol of uncertainty. $m_1\left\{\Theta\right\} \times m_2(\left\{R\right\})$ is defined to decrease degree of uncertainty.

According to the three cases of sets, the synthesis of propositions in different cases respectively from single subset, multsubset and universal set propositions discussed by $TDQBF$  are converted from uncertain propositions into certain propositions to guarantee the actuality.

The progress of combination is shown in Figure 1.

\begin{figure}[h]
\centering
\includegraphics[scale=0.6]{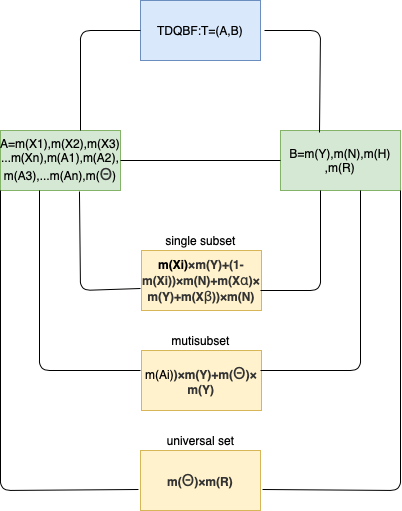}
\caption{The procedure of $TDQBF's$ combination}
\label{fig:label}
\end{figure}

\newpage

\subsection{XG Rules}

On the basis of the definition of $OQFDTI$, for describing the true situation of incident more accurately, the degree of urgency of the incident should be taken into account reasonably. For an evidence system, the weights for the same incident or proposition told by different quantum evidences are disparate to accord with the consideration about practical condition. For example, assume there exists a health assessment system, $fever$, $trauma$, $internal$ $injury$ and $healthy$ are four status of it. For this system, $OQFDTI$ can be defined as: $\Theta=\{fever, trauma, internal \ \ injury, healthy\}$. Apparently, the importance of propositions given in disparate conditions is supposed to be distinguished. For instance, in one group of evidences given during the period of $COVID-19$, $fever$ should be given more weights. Even though the quantum basic probability of assignment of $fever$ is relatively low, it is worth a great concern because it is the most central cardinal symptom of $COVID-19$. Correspondingly, the attention of $trauma$, $internal$ $injury$ and $healthy$ are supposed to be cut down. Hence, confirmation of the extent of urgency of the incidents of the evidence system is of great significance and $XG$ $Rules$ is proposed to solve the problem about how to measure the urgency of incidents. The particular process of obtaining the results from the combination of each ordinal quantum evidence can be expressed as:

(1)In accordance with $XG$ $Rules$, incidents are divided into several degrees of urgency and corresponding symbolic expressions $U^{e}$ are designed. The specific intervals of divisions are defined as:

$U^{0<e<1/2}$ considered as "Ignorable event";

$U^{1/2 \leqslant e<1}$ considered as "More or less ignorable event";

$U^{e=1}$ considered as "Normal event";

$U^{1\leqslant e<2}$ considered as "More or less urgent event";

$U^{e=2}$ considered as "Urgent event";

$U^{2\leqslant e<5/2}$ considered as "Quite urgent event";

$U^{5/2 \leqslant e <3}$ considered as "Very urgent event";

$U^{3 \leqslant e<5}$ considered as "Quite very urgent event";

$U^{5 \leqslant e}$ considered as "Very very urgent event ";

\textbf{Note: }$U^{e}$ represents the extent of importance of each piece of quantum evidence.

(2)The process of getting modified value for proposition with subscript $p$ is defined as:

\begin{equation}
MID(p)=\sum_{i=1}^{n}(U^{e} \times m_X(p))
\end{equation}

(3)The step of normalization of $MID(p)$ is defined as:
\begin{equation}
NOR(p)=\frac{MID(p)}{\sum_{p=1}^{k}	MID(p)}
\end{equation}

(4)The step of employing classical Dempster's rule of combination to combine the $NOR(p)$ for $n-1$ times is defined as:
\begin{tiny}
\begin{equation}
	Combine\quad (n-1) \ \ times\left\{
	\begin{array}{rcl}
		\begin{Bmatrix}
			NOR(1)&NOR(2)&...&NOR(p)&
		\end{Bmatrix}    \\
		\begin{Bmatrix}
			NOR(1)&NOR(2)&...&NOR(p)&
		\end{Bmatrix} \\
		.\quad .\quad .   \\
		.\quad .\quad .   \\
		.\quad .\quad .   \\
		\begin{Bmatrix}
			NOR(1)&NOR(2)&...&NOR(p)&
		\end{Bmatrix}\\
		
	\end{array} \right.
	\Longrightarrow
	\begin{Bmatrix}
		FIN(1)&FIN(2)&...&FIN(p)&
	\end{Bmatrix}
\end{equation}
\end{tiny}

\section{Application 1}
\begin{example}
	Application of virtual currency
\end{example}

Assume there exists a financial company which makes judgment about a kind of virtual currency. The $OQFDTI$ of it is given as: $\Theta=\{R,S,D,SD\}$. $R$, $S$ and $D$ represent the development of the virtual currency are $Raised$, $Steady$ and $Decreasing$ respectively. Analogously, hesitation in $Smooth$ and $Decreasing$ is indicated by $SD$. Besides, the predictions given by senors in four pieces of evidences $m_{1}$ are listed in Table \ref{data m1} and corresponding degree of reliability $m_{2}$ of the mass of $m_{1}$ is shown in Table \ref{data  m2}. Then the moments given to propositions are listed in Table \ref{data moment} and the unit of moment of the begin to forecast is seconds. Besides, modified $m_{1}$ after process with time interval are exhibited in Table \ref{data moment m1} and the value of $m_{2}$ whose hesitation is reassigned is displayed in Table \ref{data moment m2}. In addition, the initial results of combination of $TDQBF$ is enumerated in Table \ref{data TDQBF} and the degree of urgency of the incident is exhibited in Table \ref{urgency}. Finally, the comparison of final results using proposed method and traditional rule of combination in quantum field and in the form of classic probability assignment are listed respectively in Table \ref{finvalue} and Table \ref{finvalue21}. 

\begin{table}[!htbp]
\footnotesize
\centering
\caption{Quantum evidences given by  senors }
\begin{spacing}{1.80}
	\begin{tabular}{c c c c c}
		\hline
		$Evidences$ &\multicolumn{4}{c}{$Values \ \ of \ \ propositions$}\\ 
		\hline
		&$\{R\}$&$\{S\}$&$\{D\}$&$\{SD\}$\\
		$Evidence_{1}$&$0.7141 e^{0.9294j}$&$0.3317 6e^{1.0206}$&$0.4796e^{0.7399}$&$0.3873 e^{1.0965}$\\  
		&$\{SD\}$&$\{S\}$&$\{R\}$&$\{D\}$\\
		$Evidence_{2}$&$0.6481e^{0.6594j}$&$0.4899 e^{0.9376j}$&$0.5568 e^{1.0414 j}$&$0.1732e^{0.1023 j}$\\ 
		&$\{D\}$&$\{R\}$&$\{SD\}$&$\{S\}$\\
		$Evidence_{3}$&$0.3873 e^{0.9872 j}$&$0.6245 e^{0.3164j}$&$0.1732 e^{0.9260 j}$&$0.6633 e^{0.9643 j}$\\ 
		&$\{D\}$&$\{SD\}$&$\{S\}$&$\{R\}$\\
		$Evidence_{4}$&$0.7141 e^{0.2656 j}$&$0.2449 e^{1.4837j}$&$0.4000 e^{1.0661 i}$&$0.5196 e^{0.7514 j}$\\ 
		\hline
	\end{tabular}
\end{spacing}
\label{data m1}
\end{table}

\begin{table}[!htbp]\footnotesize
\centering
\caption{The value of the two-dimensional judgment  given by senors }
\begin{spacing}{1.80}
	\begin{tabular}{c c c c c}
		\hline
		$Evidences$ &\multicolumn{4}{c}{$Values \ \ of \ \ propositions$}\\ 
		\hline
		&$\{Y\}$&$\{N\}$&$\{H\}$&$\{R\}$\\
		$Evidence_{1}$&$0.5657  e^{0.9738j}$&$0.4243  e^{1.2295 j}$&$0.6245 e^{1.3758j}$&$0.3317  e^{1.2518j}$\\  
		&$\{Y\}$&$\{N\}$&$\{H\}$&$\{R\}$\\
		$Evidence_{2}$&$0.4899e^{1.2874 j}$&$0.5196  e^{0.9994 j}$&$0.5916  e^{0.8465  j}$&$0.3742 e^{1.3030  j}$\\ 
		&$\{Y\}$&$\{N\}$&$\{H\}$&$\{R\}$\\
		$Evidence_{3}$&$0.4123  e^{1.2475 j}$&$0.3742  e^{1.2780 j}$&$0.7141 e^{0.7366  j}$&$0.4243  e^{1.3231  j}$\\ 
		&$\{Y\}$&$\{N\}$&$\{H\}$&$\{R\}$\\
		$Evidence_{4}$&$0.5099 e^{1.0785 j}$&$0.4583  e^{0.6236 j}$&$0.5385  e^{1.3802 j}$&$0.4899  e^{1.1747 j}$\\ 
		\hline
	\end{tabular}
\end{spacing}
\label{data  m2}
\end{table}

\begin{table}[!htbp]\footnotesize
\centering
\caption{The value of moment of each proposition }
\begin{spacing}{1.80}
	\begin{tabular}{c c c c c}
		\hline
		$Evidences$ &\multicolumn{4}{c}{$The \ \ moment  \ \ of  \ \ occurrence$}\\ 
		\hline
		&$\{R\}$&$\{S\}$&$\{D\}$&$\{SD\}$\\
		$Evidence_{1}$&$3.1752$&$8.4246$&$40.6898$&$51.2317$\\  
		&$\{SD\}$&$\{S\}$&$\{R\}$&$\{D\}$\\
		$Evidence_{2}$&$10.2351$&$11.2819$&$15.3474$&$121.413$\\ 
		&$\{D\}$&$\{R\}$&$\{SD\}$&$\{S\}$\\
		$Evidence_{3}$&$34.7341$&$90.2663$&$150.3377$&$157.4826$\\ 
		&$\{D\}$&$\{SD\}$&$\{S\}$&$\{R\}$\\
		$Evidence_{4}$&$181.4964$&$197.2876$&$314.9728$&$528.7392$\\ 
		\hline
	\end{tabular}
\end{spacing}
\label{data moment}
\end{table}

\begin{table}[!htbp]\footnotesize
\centering
\caption{The value of first-dimensional judgment processed by time quantitative rule }
\begin{spacing}{1.80}
	\begin{tabular}{c c c c c}
		\hline
		$Evidences$ &\multicolumn{4}{c}{$Values \ \ of \ \ propositions$}\\ 
		\hline
		&$\{R\}$&$\{S\}$&$\{D\}$&$\{SD\}$\\
		$Evidence_{1}$&$0.8500   e^{0.9294 j}$&$0.1821  e^{1.0206  j}$&$0.4710   e^{0.7399   j}$&$0.1503  e^{1.0965   j}$\\  
		&$\{SD\}$&$\{S\}$&$\{R\}$&$\{D\}$\\
		$Evidence_{2}$&$0.7973 e^{0.6594  j}$&$0.5890  e^{0.9376  j}$&$0.1300   e^{1.0414  j}$&$0.0202  e^{0.1023   j}$\\ 
		&$\{D\}$&$\{R\}$&$\{SD\}$&$\{S\}$\\
		$Evidence_{3}$&$0.6224   e^{0.9872  j}$&$0.2394  e^{0.3164 j}$&$0.1944  e^{0.9260   j}$&$0.7193  e^{0.9643 j}$\\ 
		&$\{D\}$&$\{SD\}$&$\{S\}$&$\{R\}$\\
		$Evidence_{4}$&$0.9118  e^{0.2656  j}$&$0.2456  e^{1.4837  j}$&$0.2966   e^{1.0661  j}$&$0.1425  e^{0.7514  j}$\\ 
		\hline
	\end{tabular}
\end{spacing}
\label{data moment m1}
\end{table}

\begin{table}[!htbp]\footnotesize
\centering
\caption{The value of two-dimensional judgment redistributed by $DHDF$ and $QPDR $}
\begin{spacing}{1.80}
	\begin{tabular}{c c c c c}
		\hline
		$Evidences$ &\multicolumn{4}{c}{$Values \ \ of \ \ propositions$}\\ 
		\hline
		&$\{Y\}$&$\{N\}$&$\{H\}$&$\{R\}$\\
		$Evidence_{1}$&$0.8739   e^{1.1198  j}$&$0.6056 e^{1.2735   j}$&$0.1169  e^{1.3758   j}$&$0.3317  e^{1.2518   j}$\\  
		&$\{Y\}$&$\{N\}$&$\{H\}$&$\{R\}$\\
		$Evidence_{2}$&$0.6971  e^{1.1511  j}$&$0.7674  e^{0.9498   j}$&$0.1199   e^{0.8465  j}$&$0.3742  e^{1.3030   j}$\\ 
		&$\{Y\}$&$\{N\}$&$\{H\}$&$\{R\}$\\
		$Evidence_{3}$&$0.7276  e^{1.0173   j}$&$0.6305 e^{1.0474  j}$&$0.0951   e^{0.7366   j}$&$0.4243  e^{1.3231 }$\\ 
		&$\{Y\}$&$\{N\}$&$\{H\}$&$\{R\}$\\
		$Evidence_{4}$&$0.7442  e^{1.1752  j}$&$0.6150  e^{0.8433   j}$&$0.1015   e^{1.3802  j}$&$0.4899  e^{1.1747  j}$\\ 
		\hline
	\end{tabular}
\end{spacing}
\label{data moment m2}
\end{table}

\textbf{\begin{table}[!htbp]\footnotesize
	\centering
	\caption{The initial combination of result of $TDQBF$}
	\begin{spacing}{1.80}
		\begin{tabular}{c c c c c}
			\hline
			$Evidences$ &\multicolumn{4}{c}{$Values \ \ of \ \ propositions$}\\ 
			\hline
			&$\{R\}$&$\{S\}$&$\{D\}$&$\{SD\}$\\
			$Evidence_{1}$&$0.8898    e^{1.4871 j}$&$0.7403 e^{1.4550   j}$&$0.8273   e^{1.4456    j}$&$0.1314  e^{2.2162    j}$\\  
			&$\{R\}$&$\{S\}$&$\{D\}$&$\{SD\}$\\
			$Evidence_{2}$&$1.2856   e^{1.2455   j}$&$1.1201   e^{1.3439  j}$&$1.2049    e^{1.3095  j}$&$0.5559  e^{1.8105  j}$\\ 
			&$\{R\}$&$\{S\}$&$\{D\}$&$\{SD\}$\\
			$Evidence_{3}$&$0.7347   e^{1.1842j}$&$0.7877   e^{1.2503 j}$&$0.7787   e^{1.2455 j}$&$0.1415  e^{1.9433j }$\\ 
			&$\{R\}$&$\{S\}$&$\{D\}$&$\{SD\}$\\
			$Evidence_{4}$&$0.6409    e^{1.1351 j}$&$0.5711   e^{1.2676    j}$&$0.7200 e^{1.4523 j}$&$0.1828  e^{2.6589  j}$\\ 
			\hline
		\end{tabular}
	\end{spacing}
	\label{data TDQBF}
\end{table}}

\begin{table}[h]\footnotesize
\centering
\caption{The degree of urgency of each piece of evidence $U^{e}$ given by $XG \ \ Rules$}
\begin{spacing}{1.80}
	\begin{tabular}{c c c c c  }\hline
		
		$Quantum \ \ evidence $	& $Evidence1$ & $Evidence2$ & $Evidence3$ & $Evidence4$  \\  \hline
		$Degree$ $of$ $ergency$ & $U^{3.1741}$& $U^{1.3627}$ & $U^{1.0183}$ & $U^{2.7549}$ \\ \hline
	\end{tabular}
\end{spacing}
\label{urgency}
\end{table}

\

\begin{table}[h]\tiny
\centering
\caption{The comparison of results combined using proposed method and traditional rule of combination in quantum field }
\begin{spacing}{1.80}
	\begin{tabular}{c c c c c }\hline
		
		$Proposition$ 	& $R$ & $S$ & $D$ & $SD$ \\  \hline
		$The \ \ improved \ \ combined \ \ values $ & $0.3563   e^{-1.4645 j}$& $0.5375   e^{-0.6216   j}$ & $0.7643   e^{-0.5740 j}$ & $0.0016 e^{1.6883j}$  \\ \hline
		$Proposition $&  $R$ & $S$ & $D$ & $SD$  \\  \hline
		$Combined \ \ values $& $0.2393e^{-2.6170j}$& $0.7754e^{-1.6379j}$& $0.5840e^{-2.7286j}$& $0.0195e^{-1.480j}$ \\ \hline
	\end{tabular}
\end{spacing}
\label{finvalue}
\label{}

\end{table}

\begin{table}[h]\footnotesize
\centering
\caption{The comparison of results combined using proposed method and traditional rule of combination in the form of classic probability assignment }
\begin{spacing}{1.80}
	\begin{tabular}{c c c c c }\hline
		
		$Proposition  $& $R$ & $S$ & $D$ & $SD$  \\  \hline
		$The \ \ improved \ \ combined \ \ values $& $0.1270$& $0.2889$ & $0.5841$ & $2.57786E-06$ \\ \hline
		$Proposition  $& $R$ & $S$ & $D$ & $SD$ \\  \hline
		$Combined \ \ values $& $0.0572$ & $0.6013$ &$ 0.3410$ &$ 0.0003$ \\ \hline
	\end{tabular}
\end{spacing}
\label{finvalue21}
\end{table}

\newpage
By comparing and analyzing the Table \ref{finvalue21}, clear conclusion can be obtained. Traditional $OQFOD$ can process elementary predictive parsing. However, traditional frame does not take the closeness of degree of any two sequential propositions and the urgency of incidents into consideration which cause the outcomes of prediction which does not have higher accuracy and accords with the situation of true world. The quantum probability assignment of proposition $S$ told by proposed method goes down dramatically which manifests that proposed method exists a  clearer judgment. Besides, the quantum probability assignment of proposition $D$ increases which  illustrates that the development of this virtual currency is with a high possibility to decline under the consideration of all influence factors. In addition, the basic probability assignment of $SD$ and $RSD$ are so small that it can be ignored, which also illustrates that proposed method is extremely advantageous in eliminating uncertainty and provide explicit indicator of the true situation. The reason to observe such a big difference in the results obtained by the proposed and traditional method is that the value of quantum basic probability assignment of mutisubset proposition is allocated suitably to single subset proposition to reduce the uncertainty contained in $OQFDTI$. What's more, the degree of hesitation of the dual judgment system is also distributed to the membership and non-membership in a similar way to reflect information exerted by multi-source senors. Hence, the prominent positive performance is possessed by  method proposed.

\newpage

\section{Application 2}
\begin{example}
Application of meteorological disaster prediction
\end{example}
How to make correct precise judgment to meteorological disaster for preventing a large amount of economic losses is still an urgent issue. Assume there exists a quantum frame of discernment about meteorological disaster which is denoted as $\Theta$=$\left\{H,D,S,HS,HDS\right\}$. The quantum evidences under the quantum frame of discernment are given in Table \ref{data m1_2}. Besides, the belief of propositions are listed in Table \ref{data m2_2}. And the moment given of each propositions of quantum evidences is shown in Table \ref{data moment application2}. Then the more precise data processed by $Time \ \ quantitative \ \ rule $ is shown in Table \ref{data moment m1_2} and \ref{data  moment m2_2}, and the results combined using them by $TDQBF$ are listed in Table \ref{data $TBQDF$ application2}. The quantum and classic probability form of comparison of final results are represented in Table \ref{finvalue2 Euler} and \ref{finvalue2} respectively.

\begin{table}[!htbp]
\footnotesize
\centering
\caption{Quantum evidences given by  senors}
\begin{spacing}{1.80}
	\begin{tabular}{c c c c c c}
		\hline
		$Evidences$ &\multicolumn{5}{c}{$Values \ \ of \ \ propositions$}\\ 
		\hline
		&$\{H\}$&$\{D\}$&$\{S\}$&$\{HS\}$&$\{HDS\}$\\
		$Evidence_{1}$&$0.6083  e^{0.8916 j}$&$0.3742  e^{1.1948 j}$&$0.5292 e^{1.1009 j}$&$0.2828  e^{1.5301 j}$&$0.3606  e^{1.2776  j}$\\  
		&$\{D\}$&$\{HDS\}$&$\{HS\}$&$\{S\}$&$\{H\}$\\
		$Evidence_{2}$&$0.2000 e^{1.4842 j}$&$0.4123  e^{1.2508 j}$&$0.5099  e^{1.0976  j}$&$0.2646 e^{1.4625 j}$&$0.6782   e^{0.8703   j}$\\ 
		&$\{HDS\}$&$\{D\}$&$\{S\}$&$\{HS\}$&$\{H\}$\\
		$Evidence_{3}$&$0.2449 e^{1.5099 j}$&$0.3317  e^{1.1739 j}$&$0.3742  e^{1.2794  j}$&$0.4123  e^{1.1101  j}$&$0.7211   e^{0.5107  j}$\\ 
		&$\{S\}$&$\{HDS\}$&$\{D\}$&$\{HS\}$&$\{H\}$\\
		$Evidence_{4}$&$0.5831  e^{1.1545  j}$&$0.4243  e^{1.2250 j}$&$0.4359  e^{1.2319 j}$&$0.1414  e^{1.4310 j}$&$0.5196   e^{0.9987 j}$\\ 
		&$\{S\}$&$\{HS\}$&$\{H\}$&$\{HDS\}$&$\{D\}$\\
		$Evidence_{5}$&$0.1414   e^{1.2899   j}$&$0.1732  e^{1.4951  j}$&$0.5657   e^{0.8162  j}$&$0.5916  e^{1.0703  j}$&$0.5292    e^{1.0643 j}$\\ 
		\hline
	\end{tabular}
\end{spacing}
\label{data m1_2}
\end{table}

\begin{table}[!htbp]
\footnotesize
\centering
\caption{     The value of the two-dimensional judgment  given by senors}
\begin{spacing}{1.80}
	\begin{tabular}{c c c c c }
		\hline
		$Evidences$ &\multicolumn{4}{c}{$Values \ \ of \ \ propositions$}\\ 
		\hline
		&$\{Y\}$&$\{N\}$&$\{H\}$&$\{R\}$\\
		$Evidence_{1}$&$0.6083  e^{0.8769 }$&$0.3317   e^{1.2103  j}$&$0.5385  e^{1.1151  j}$&$0.4796  e^{1.1314  j}$\\  
		&$\{Y\}$&$\{N\}$&$\{H\}$&$\{R\}$\\
		$Evidence_{2}$&$0.6481  e^{0.8308  j}$&$0.4243  e^{1.2044  j}$&$0.3742   e^{1.0659   j}$&$0.5099  e^{0.9490  j}$\\ 
		&$\{Y\}$&$\{N\}$&$\{H\}$&$\{R\}$\\
		$Evidence_{3}$&$0.4583  e^{1.0399  j}$&$0.5831  e^{0.9960  j}$&$0.5745   e^{0.9244  j}$&$0.3464   e^{1.2263  j}$\\ 
		&$\{Y\}$&$\{N\}$&$\{H\}$&$\{R\}$\\
		$Evidence_{4}$&$0.7280   e^{0.6283   j}$&$0.3606  e^{1.2637  j}$&$0.4359  e^{1.1018  j}$&$0.3873  e^{1.0191  j}$\\ 
		&$\{Y\}$&$\{N\}$&$\{H\}$&$\{R\}$\\
		$Evidence_{5}$&$0.7810    e^{0.6170    j}$&$0.3873  e^{0.9885  j}$&$0.4123   e^{1.1082   j}$&$0.2646  e^{1.4800   j}$\\ 
		\hline
	\end{tabular}
\end{spacing}
\label{data m2_2}
\end{table}

\begin{table}[!htbp]\footnotesize
\centering
\caption{The value of moment of each proposition}
\begin{spacing}{1.80}
	\begin{tabular}{c c c c c c}
		\hline
		$Evidences$ &\multicolumn{5}{c}{$The \ \ moment  \ \ of  \ \ occurrence$}\\ 
		\hline
		&$\{H\}$&$\{D\}$&$\{S\}$&$\{HS\}$&$\{HDS\}$\\
		$Evidence_{1}$&$17.9022$&$58.4711$&$162.4293$&$174.1377$&$199.3269$\\  
		&$\{D\}$&$\{HDS\}$&$\{HS\}$&$\{S\}$&$\{H\}$\\
		$Evidence_{2}$&$41.3827$&$152.5734$&$244.3795$&$377.9305$&$445.1812$\\  
		&$\{HDS\}$&$\{D\}$&$\{S\}$&$\{HS\}$&$\{H\}$\\
		$Evidence_{3}$&$159.0479$&$177.2146$
		&$194.4576$&$247.5243$&$307.2954$\\  
		&$\{S\}$&$\{HDS\}$&$\{D\}$&$\{HS\}$&$\{H\}$\\
		$Evidence_{4}$&$1047.2412$&$1372.5814$&$1779.4607$&$1993.2938$&$2451.3926$\\  
		&$\{S\}$&$\{HS\}$&$\{H\}$&$\{HDS\}$&$\{D\}$\\
		$Evidence_{5}$&$1973.4528$&$1324.5271$&$1473.3968$&$1709.4353$&$1877.2964$\\  
		\hline
	\end{tabular}
\end{spacing}
\label{data moment application2}
\end{table}

\begin{table}[!htbp]
\footnotesize
\centering
\caption{The value of first-dimensional judgment processed by time quantitative rule}
\begin{spacing}{1.80}
	\begin{tabular}{c c c c c c}
		\hline
		$Evidences$ &\multicolumn{5}{c}{$Values \ \ of \ \ propositions$}\\ 
		\hline
		&$\{H\}$&$\{D\}$&$\{S\}$&$\{HS\}$&$\{HDS\}$\\
		$Evidence_{1}$&$0.8755   e^{0.8916  }$&$0.1996   e^{1.1948   j}$&$0.1956  e^{1.1009   j}$&$0.3077  e^{1.5301   j}$&$0.2463   e^{1.2776   j}$\\  
		&$\{D\}$&$\{HDS\}$&$\{HS\}$&$\{S\}$&$\{H\}$\\
		$Evidence_{2}$&$0.3673  e^{1.4842 j}$&$0.3106  e^{1.2508 j}$&$0.5373   e^{1.0976   j}$&$0.2319  e^{1.4625 j}$&$0.6528   e^{0.8703   j}$\\ 
		&$\{HDS\}$&$\{D\}$&$\{S\}$&$\{HS\}$&$\{H\}$\\
		$Evidence_{3}$&$0.5026 ^{1.5099 j}$&$0.4333  e^{1.1739 j}$&$0.4595   e^{1.2794  j}$&$0.2867   e^{1.1101  j}$&$0.5160    e^{0.5107  j}$\\ 
		&$\{S\}$&$\{HDS\}$&$\{D\}$&$\{HS\}$&$\{H\}$\\
		$Evidence_{4}$&$0.8280   e^{1.1545  j}$&$0.3441   e^{1.2250 j}$&$0.2416   e^{1.2319 j}$&$0.1081  e^{1.4310 j}$&$0.3550   e^{0.9987 j}$\\ 
		&$\{S\}$&$\{HS\}$&$\{H\}$&$\{HDS\}$&$\{D\}$\\
		$Evidence_{5}$&$0.2694   e^{1.2899   j}$&$0.1423   e^{1.4951  j}$&$0.5506    e^{0.8162  j}$&$0.4576   e^{1.0703  j}$&$0.6282     e^{1.0643 j}$\\ 
		
		\hline
	\end{tabular}
\end{spacing}
\label{data moment m1_2}
\end{table}

\begin{table}[!htbp]
\footnotesize
\centering
\caption{The value of two-dimensional judgment redistributed by $DHDF $ and $QPDR$ }
\begin{spacing}{1.80}
	\begin{tabular}{c c c c c }
		\hline
		$Evidences$ &\multicolumn{4}{c}{$Values \ \ of \ \ propositions$}\\ 
		\hline
		&$\{Y\}$&$\{N\}$&$\{H\}$&$\{R\}$\\
		$Evidence_{1}$&$0.9568   e^{0.9646  }$&$0.4368  e^{1.1873  j}$&$0.0781   e^{1.1151   j}$&$0.4796  e^{1.1314  j}$\\  
		&$\{Y\}$&$\{N\}$&$\{H\}$&$\{R\}$\\
		$Evidence_{2}$&$0.8529  e^{0.8879   j}$&$0.5132  e^{1.1803    j}$&$0.0753  e^{1.0659 j}$&$0.5099  e^{0.9490  j}$\\
		&$\{Y\}$&$\{N\}$&$\{H\}$&$\{R\}$\\
		$Evidence_{3}$&$0.6315   e^{1.0082  j}$&$0.8645  e^{0.9727    j}$&$0.1184    e^{0.9244   j}$&$0.3464   e^{1.2263  j}$\\ 
		&$\{Y\}$&$\{N\}$&$\{H\}$&$\{R\}$\\
		$Evidence_{4}$&$0.9945     e^{0.7614   j}$&$0.4308  e^{1.2371    j}$&$0.0753   e^{1.1018  j}$&$0.3873  e^{1.0191  j}$\\ 
		&$\{Y\}$&$\{N\}$&$\{H\}$&$\{R\}$\\
		$Evidence_{5}$&$1.0240      e^{0.7402  j}$&$0.4525  e^{1.0058  j}$&$0.0800    e^{1.1082   j}$&$0.2646  e^{1.4800   j}$\\ 
		
		\hline
	\end{tabular}
\end{spacing}
\label{data  moment m2_2}
\end{table}

\begin{table}[!htbp]
\footnotesize
\centering
\caption{The initial combination of result of $TDQBF$}
\begin{spacing}{1.80}
	\begin{tabular}{c c c c c c}
		\hline
		$Evidences$ &\multicolumn{5}{c}{$Values \ \ of \ \ propositions$}\\ 
		\hline
		&$\{H\}$&$\{D\}$&$\{S\}$&$\{HS\}$&$\{HDS\}$\\
		$Evidence_{1}$&$1.2655   e^{1.7841    }$&$0.6325    e^{1.7068    j}$&$0.9030    e^{1.8523     j}$&$0.5260    e^{2.3825    j}$&$0.1181     e^{2.4090    j}$\\  
		&$\{D\}$&$\{HDS\}$&$\{HS\}$&$\{S\}$&$\{H\}$\\
		$Evidence_{2}$&$1.3695   e^{1.6299  j}$&$0.9338  e^{1.7453 j} $&$1.2125    e^{1.7084  j}$&$0.7212  e^{2.0416    j}$&$0.1584  e^{2.1998   j}$\\ 
		&$\{HDS\}$&$\{D\}$&$\{S\}$&$\{HS\}$&$\{H\}$\\
		$Evidence_{3}$&$0.9973   ^{1.3551   j}$&$1.1733  e^{1.3923   j}$&$1.0459     e^{1.2848   j}$&$0.4545   e^{2.1567  j}$&$0.1501      e^{2.4002    j}$\\ 
		&$\{S\}$&$\{HDS\}$&$\{D\}$&$\{HS\}$&$\{H\}$\\
		$Evidence_{4}$&$1.0344     e^{1.6015    j}$&$0.6765  e^{1.6193   j}$&$1.3415   e^{1.6284  j}$&$0.4480  e^{2.0356  j}$&$0.1333   e^{2.2441  j}$\\ 
		&$\{S\}$&$\{HS\}$&$\{H\}$&$\{HDS\}$&$\{D\}$\\
		$Evidence_{5}$&$1.4469     e^{1.5199     j}$&$1.0472     e^{1.5310  j}$&$1.2791  e^{1.6005  j}$&$0.7779   e^{1.8828   j}$&$0.1662     e^{2.5443  j}$\\ 
		\hline
	\end{tabular}
\end{spacing}
\label{data $TBQDF$ application2}
\end{table}

\begin{table}[h]\footnotesize
\centering
\caption{The degree of urgency of each piece of evidence $U^{e}$ given by $XG \ \ Rules$}
\begin{spacing}{1.80}
	\begin{tabular}{c c c c c c }\hline
		
		$Quantum \ \ evidence $	& $Evidence1$ & $Evidence2$ & $Evidence3$ & $Evidence4$ &$Evidence5$\\  \hline
		$Degree$ $of$ $ergency$ & $U^{2.4342}$& $U^{1.5325}$ & $U^{4.1342}$ & $U^{1.0537}$ &$U^{0.8321}$ \\ \hline
	\end{tabular}
\end{spacing}
\label{urgency application2}
\end{table}

\begin{table}[h]\tiny
\centering
\caption{The comparison of results combined using proposed method and traditional rule of combination in quantum field }
\begin{spacing}{1.80}
	\begin{tabular}{c c c c c c}\hline
		
		$Proposition$ 	&$\{H\}$&$\{D\}$&$\{S\}$&$\{HS\}$&$\{HDS\}$\\  \hline
		$The \ \ improved \ \ combined \ \ values $ &$0.7810  e^{2.6678    }$&$0.0428   e^{2.0078    j}$&$0.6230   e^{2.7456    j}$&$0.0066  e^{-1.7475    j}$&$0.0000   e^{-0.7723     j}$\\ \hline
		$Proposition $	&$\{H\}$&$\{D\}$&$\{S\}$&$\{HS\}$&$\{HDS\}$\\  \hline
		$Combined \ \ values $& $0.9320 e^{-2.1675 j}$& $0.0664 e^{-1.1650 j}$& $0.3538 e^{-1.1911 j}$& $0.0426 e^{-1.0823 j}$  & $0.0025 e^{-0.9505 j}$\\ \hline
	\end{tabular}
\end{spacing}
\label{finvalue2 Euler}
\label{}
\end{table}

\begin{table}[h]\footnotesize
\centering
\caption{The comparison of results combined using proposed method and traditional rule of combination in the form of classic probability assignment}
\begin{spacing}{1.80}
	\begin{tabular}{c c c c c c}\hline
		
		$Proposition $	   &$\{H\}$&$\{D\}$&$\{S\}$&$\{HS\}$&$\{HDS\}$ \\  \hline
		$The \ \ improved \ \ combined \ \ values $	$Values$ & $0.6100  $& $0.0018  $ & $0.3882  $ & $0.0000 $ &$ 0.0000  $\   \\ \hline
		$Proposition $   &$\{H\}$&$\{D\}$&$\{S\}$&$\{HS\}$&$\{HDS\}$ \\  \hline
		$Combined \ \ values $& $0.8686 $ & $0.0044 $ &$ 0.1252 $ &$ 0.0018 $ &$0.0000  $ \\ \hline
	\end{tabular}
\end{spacing}
\label{finvalue2}
\end{table}

\newpage

The advantages of the method in this paper can be represented by taking belief of propositions, time interval and urgency of evidence into consideration in this paper. The data in Table \ref{finvalue2 Euler} disposed by new method proposed in this paper can be analyzed that the probability of hail occurring occupies the biggest part of the whole probability distribution of meteorological disaster in this city, and then drought and dusk storms almost take up the rest of probability, at last it is difficult to observe the phenomenon of the occurrence of the propositions. And the data in Table \ref{finvalue2} managed by traditional method can be read that hail is the most possible to occur and almost nothing else happened. In the actual life, more than one meteorological disasters can occur in a similar period time. So, the prediction shown by traditional methods is against real life and the superiority of method mentioned in this paper can be fully demonstrated for the non-uniqueness of the possibly predictive disasters. The concrete preponderances can be discovered in Table \ref{finvalue2 Euler} and \ref{finvalue2}.

\section{Conclusion}

In this paper, a completely new method in alleviating uncertainties in quantum information is proposed. To achieve the established goal, a dual check system and a rule revised for expert system are designed. The given applications proves that the proposed method successfully extracts truly useful information from quantum evidences. Compared with the traditional method, the proposed method reduces the degree of hesitancy and produce clear judgment on the current situations. All in all, the proposed method offers a completely new vision on the uncertainty management of quantum information and can produce intuitive and reasonable results.

\section*{Acknowledgment}
The authors greatly appreciate the reviews' suggestions and the editor's encouragement. This research is supported by the National Natural Science Foundation
 of China (No. 62003280).
\bibliographystyle{elsarticle-num}
\bibliography{cite}
\end{document}